%% file: main.tex
\documentclass[11pt]{article}
\usepackage[preprint]{acl}    %
\usepackage{times}
\usepackage{latexsym}
\usepackage[T1]{fontenc}
\usepackage[utf8]{inputenc}
\usepackage{microtype}
\usepackage{booktabs}
\usepackage{array}
\usepackage{amsmath}
\usepackage{graphicx}
\usepackage{listings}
\usepackage{tikz}
\usetikzlibrary{positioning,arrows.meta,fit} 

\usepackage{xcolor}
\usepackage{placeins}

\title{Instrument-Anchored LLM Judges Transfer Expert\\
Counseling-Quality Ratings Across Three German Corpora}

\author{Tobias Hallmen \\
  Chair for Human-Centered \\
  Artificial Intelligence \\
  University of Augsburg \\
  \texttt{tobias.hallmen@uni-a.de}
  \And
  Elisabeth Andr\'e \\
  Chair for Human-Centered \\
  Artificial Intelligence \\
  University of Augsburg \\
  \texttt{elisabeth.andre@uni-a.de}}

\begin{document}
\maketitle

\begin{abstract}
Automatic assessment of communication quality in counseling conversations is bottlenecked by data: expert-rated corpora are small and expensive to grow. Therefore we study cross-domain prediction of the expert's overall impression across three German simulated-counseling corpora from one university (two general-practice medical, one parent--teacher; $n{=}195$ sessions). Features are construct scores that four small LLM judges assign from the transcript, drawn from the experts' own rating instruments, plus a compact nonverbal block; all models are open-weight and run locally, preserving data protection. Trained on the other two corpora, this stack predicts the held-out corpus at Spearman $\rho = .533$; in-domain training on the same sessions falls about a fifth short ($.420$; paired gap $+.113$, 95\% CI $[+.067, +.157]$), and the gap holds at matched training size and on identical test sessions. The signal lies in what is said: the four judges alone transfer at $.523$, and the nonverbal block adds $+.01$ (CI includes zero). Asking a judge for instrument constructs instead of generic dialogue qualities raises its transfer by about 30\% ($.32$ to $.42$); restoring speaker attribution where per-speaker audio was lost is worth $+.08$ to $+.10$ on that fold. Where a program's own ratings are scarce, other programs' data is a viable training set and language carries it.
\end{abstract}

\section{Introduction}

Professional communication training programs in medicine and education rely on simulated conversations, such as objective structured clinical examinations (OSCEs) and parent--teacher role-plays, that are rated by human experts on global communication quality. Such ratings are expensive: they consume scarce expert time, and professional actors where simulations use them, and each such program produces only dozens of sessions, rated with program-specific instruments. Automating them would enable formative feedback at scale, but the obvious supervised route collides with the data reality: at $53$--$85$ sessions per corpus, within-domain models stay well below the reliability reference of the ratings themselves (Section~\ref{sec:discussion}).

\begin{figure*}[t]
\centering
\resizebox{\textwidth}{!}{%
\begin{tikzpicture}[
  font=\small,
  box/.style={draw=black!55, rounded corners=2pt, align=center,
              inner sep=4pt, minimum height=8mm},
  data/.style={box, fill=black!6},
  proc/.style={box, fill=blue!8},
  instr/.style={box, fill=green!9, draw=green!40!black!60},
  arr/.style={-{Stealth[length=2mm]}, black!60}]
\node[data] (corpora) {3 corpora\\[-1pt]\scriptsize \textsc{Ticks} 57 $\cdot$ \textsc{Vacc} 53 $\cdot$ \textsc{Teach} 85};
\node[data, right=7mm of corpora, yshift=9mm] (tr) {two-speaker transcript\\[-1pt]\scriptsize time-ordered, labels B:/G:};
\node[data, right=7mm of corpora, yshift=-9mm] (av) {audio + video\\[-1pt]\scriptsize per speaker};
\node[proc, right=7mm of tr] (judges) {LLM judges $\times 4$\\[-1pt]\scriptsize construct scores 1--5 (JSON)};
\node[instr, above=3.5mm of judges] (instr) {expert instruments\\[-1pt]\scriptsize BGR $\cdot$ OSCE checklist};
\node[proc, right=7mm of av] (nv) {nonverbal features\\[-1pt]\scriptsize face/voice affect, gaze/smile,\\[-2pt]\scriptsize speech rate, dyadic $\Delta$};
\node[proc, right=7mm of judges, yshift=-9mm] (z) {feature\\[-1pt]standardization\\[-1pt]\scriptsize training statistics};
\node[proc, right=7mm of z] (model) {PCA +\\[-1pt]elastic net\\[-1pt]\scriptsize train: 2 pooled domains};
\node[data, right=7mm of model] (out) {expert overall\\[-1pt]rating 1--5\\[-1pt]\scriptsize held-out domain};
\draw[arr] (corpora.east) -- ++(2.5mm,0) |- (tr.west);
\draw[arr] (corpora.east) -- ++(2.5mm,0) |- (av.west);
\draw[arr] (tr) -- (judges);
\draw[arr] (av) -- (nv);
\draw[arr] (instr) -- node[right=1pt]{\scriptsize constructs} (judges);
\draw[arr] (judges.east) -- ++(2.5mm,0) |- (z.west);
\draw[arr] (nv.east) -- ++(2.5mm,0) |- (z.west);
\draw[arr] (z) -- (model);
\draw[arr] (model) -- (out);
\end{tikzpicture}}
\caption{Pipeline over the three corpora (Section~\ref{sec:data}): LLM judges score generic dialogue qualities and instrument-derived constructs from both
speakers' transcript, compact nonverbal features come from the recordings,
all blocks are standardized with training statistics, and a model trained on
two domains predicts the third's expert rating.}
\label{fig:pipeline}
\end{figure*}
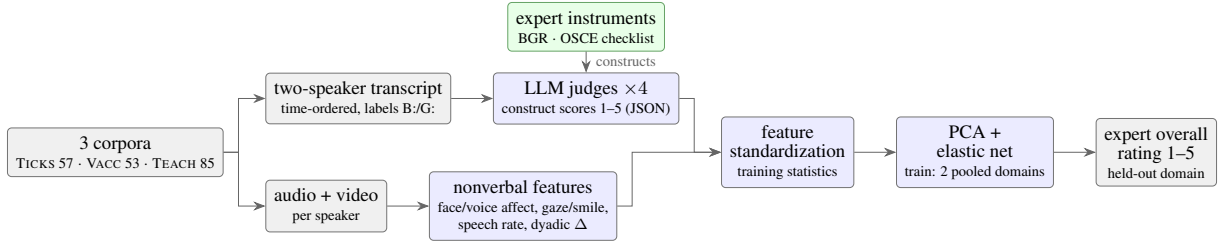

This work asks whether the assessment itself transfers: can a model trained on other counseling domains predict the experts' overall impression in an unseen one? We use three German-language corpora of simulated dyadic counseling, each pairing a student with a simulated interlocutor: general-practice consultations about tick bites, vaccination counseling, and parent--teacher conversations, each rated by domain experts on a 1--5 overall-impression scale. The programs share one university, and their three instruments start from different counseling theories but operationalize overlapping techniques and end in the same kind of global rating (Section~\ref{sec:data}), which makes transfer plausible. We evaluate leave-one-domain-out (LODO), so the test domain contributes no training sessions, and compare \emph{cross-domain} training on the other two corpora with \emph{in-domain} cross-validation inside the held-out corpus, on the same sessions.

Our primary system is the \emph{compact stack}: construct scores from four small open-weight large language models (LLM) judges reading the two-speaker transcript, plus a compact nonverbal block, with the judge membership chosen by nested selection on the training corpora alone (Section~\ref{sec:protocol}). Two of the judges ask constructs derived from the instruments the experts actually used, the Berlin Global Rating \citep{scheffer2009validierung} and an OSCE checklist of empathic techniques, so the scores also name what a learner would practice. Trained on the other two corpora, the stack predicts the held-out corpus at $\rho = .53$, against $.42$ when trained inside it; the four judges alone transfer at $.52$.

\paragraph{Contributions.} (1) \textbf{Cross-domain training for expert overall ratings:} a leave-one-domain-out comparison with in-domain training on the same sessions, across three counseling corpora from two disciplines, in which in-domain training falls about a fifth short of training on the other programs (Section~\ref{sec:results}). (2) \textbf{Instrument-derived LLM judges as features:} small open-weight judges score the constructs of the experts' own instruments from the transcript; asking a judge for the experts' instrument constructs instead of generic dialogue qualities raises its transfer by about 30\%, and judges from three model families ensemble above any single one (Section~\ref{sec:results}). (3) \textbf{What speaker attribution is worth:} for the corpus that lost its per-role audio, speaker-attribution error priced in predictive validity rather than in diarization error rate (Section~\ref{sec:repair}).

\section{Related Work}
\label{sec:related}
Prior studies of these corpora are correlational or single-domain. They relate expert ratings of the medical consultations to nonverbal behavior, arousal and smiling \citep{zerbini2025multimeasure,schneider2025smiling}, cues we test as predictors across domains; feed extracted features back to trainees \citep{gietl2026kifeedback,gietl2026foerderung,hallmen2025aibasedfeedbackcounsellingcompetence}, describing behavior where we predict the expert's rating; and annotate the phases and techniques of a counseling phase model \citep{benien2003schwierige}, where encoder baselines degrade across domains \citep{hallmen-etal-2026-annotating}, labels we turn into judge constructs for a session-level rating.

LLM-as-judge is established for machine-generated text \citep{zheng2023judging,liu2023geval} but biased and inconsistent \citep{wang2023fair,stureborg2024inconsistent,bavaresco2025llms,panickssery2024llm}. We instead use judges as feature extractors over \emph{human} behavior \citep{ziems2024can,rathje2024gpt,withanagedon2024phases,withanagedon2026mllm}: small local open-weight models (for privacy) score constructs from validated instruments without observing expert ratings; the BGR+OSCE battery includes an overall-impression construct, and a separate regressor maps all scores to the rating. Psychotherapy work predicts codes and competence from transcripts \citep{xiao2015rate,tanana2016comparison,gibson2016deep,gibson2022multilabel,flemotomos2022automated,flemotomos2021automated,perezrosas2019counselor,eberhardt2025llmscales,andre2026ppmp} in-domain, with hundreds to thousands of sessions; we test on an unseen domain with 53--85 sessions each. The public counseling corpora carry utterance-level codes or a binary quality label \citep{wu2022annomi,perezrosas2019counselor}, not an expert overall rating of a simulated session, so they cannot enter our comparison. \citet{schwartz2024multimodal} combine modalities with shallow language features, which our judges replace with construct-level scores; classroom scoring stays within one instrument \citep{wang2023chatgpt,ramakrishnan2023class}, while we transfer across three. Text dominates multimodal impressions \citep{naim2018interview,hazarika2020misa,gkoumas2021what,hallmen2025semantic}; we measure what the nonverbal channel adds under domain shift. Models degrade across corpora, which adversarial domain generalization counters \citep{gideon2021addog}; we add no such step and rely on the construct lineage the three instruments share (Section~\ref{sec:data}).

\section{Data}
\label{sec:data}

The data are three German-language corpora of simulated dyadic counseling conversations, recorded in communication training programs at one German university. In each session one \emph{assessed} person (a medical student or a teacher-education student) converses with a \emph{simulated interlocutor} (a trained actor, or a peer in role). All three collections were designed for per-speaker capture, one camera on each participant's face and one microphone per participant, so that role attribution follows from the channel rather than from a model.

\textbf{\textsc{Ticks}} (57 sessions): medical students hold a simulated general-practice consultation in which a standardized patient worries about a tick bite she has already removed, embedded in a longitudinal communication curriculum \citep{zerbini2024komcua}; one expert rated all sessions with the Berlin Global Rating (BGR) \citep{scheffer2009validierung}, a second expert a subset \citep{zerbini2025multimeasure}. \textbf{\textsc{Vacc}} (53 sessions): medical students counsel a trained actor playing a parent who hesitates about the scheduled vaccination of their three-month-old infant; four communication experts split the sessions and rated them with the BGR \citep{schneider2025smiling}. In \textsc{Vacc} the per-role audio was lost to a lapel-microphone failure, so both speakers share one mixed channel: speaker labels come from diarizing it, repaired against human speech-activity annotation (Section~\ref{sec:repair}), and overlapping speech cannot be split by speaker, so overlap features (who talks over whom) are zero by construction. Its per-role video is unaffected. \textbf{\textsc{Teach}} (85 sessions): teacher-education students in a counseling-competence seminar hold school-related parent--teacher conversations, as peer role-plays and as final simulations with professional actors \citep{hallmen-etal-2026-annotating,hallmen2025aibasedfeedbackcounsellingcompetence}; each session was rated once by one of four school counseling experts on the questionnaire's overall item.

\textsc{Ticks} and \textsc{Vacc} are the closer pair: medical students, the BGR instrument, a doctor--patient dyad and a rater pool overlapping by one person, though topic and recording differ. \textsc{Teach} shares the simulated-counseling format, but not the instrument, the rater pool or the dyad type. For all three corpora we include every session with a complete audio-video recording and an available expert overall rating at analysis time; session counts therefore differ from the analysis subsets of the corpus publications, which also report the participant demographics.

\begin{table}[t]
\centering\small
\begin{tabular}{lccc}
\toprule
 & \textsc{Ticks} & \textsc{Vacc} & \textsc{Teach} \\
\midrule
sessions & 57 & 53 & 85 \\
duration (min) & 7.1{\tiny$\pm$0.6} & 6.9{\tiny$\pm$0.9} & 9.5{\tiny$\pm$2.3} \\
assessed role & doctor & doctor & teacher \\
interlocutor role & patient & patient & parent \\
target instrument & BGR & BGR & global item \\
rating & 1--5 & 1--5 & 1--5 / 1--4 \\
expert raters & 2 & 4 & 4 \\
agreement (ICC) & .80 & .77--.85 & .53--.82 \\
\bottomrule
\end{tabular}
\caption{Corpora. BGR = Berlin Global Rating overall impression (inverted,
5 = best). ICC = between-expert agreement on the double-rated subsets
(\textsc{Ticks}/\textsc{Vacc}) and single- to average-rater ICC on the
four-rater panel (\textsc{Teach}, parallel item; Appendix~\ref{app:raters}).
\textsc{Teach} switched its item from five to four points after 24 sessions
(Appendix~\ref{app:equating}).}
\label{tab:data}
\end{table}

The prediction target is one expert's overall rating of the assessed person per session, on five levels from 5 = very good to 1 = very bad: the BGR overall item (\emph{Gesamteindruck}, inverted) for the two medical corpora and the global item (\emph{Gesamturteil}) of the teacher questionnaire. After its first 24 sessions \textsc{Teach} used a four-point version of that item, whose ratings we re-spread onto the five-point scale (Appendix~\ref{app:equating}), as they re-rated the 24 also on 1--4. \textsc{Vacc} and \textsc{Teach} distribute primary ratings across four experts each, one rater per session, and additional ratings exist on overlap subsets, with pairwise exact agreement of $.42$--$.47$ (Appendix~\ref{app:raters}): the targets are single-rater scores carrying one expert's measurement error, which the reliability reference of Section~\ref{sec:discussion} rests on. Both medical corpora also contain the standardized patient's and the student's own BGR ratings, which we use to estimate human cross-perspective agreement (Appendix~\ref{app:constructs}). Most ratings lie at 3--4, as expected from a student population, which motivates rank over absolute-error metrics (Section~\ref{sec:protocol}).

All three are counseling conversations whose instruments stem from different theories: client-centered active listening and empathic techniques such as NURSE (Naming, Understanding, Respecting, Supporting, Exploring) \citep{back2005nurse} for the medical corpora, a counseling-phase model and conversation techniques \citep{benien2003schwierige} for \textsc{Teach}. Their operationalizations converge: the teacher questionnaire's subscales mirror the BGR's item structure, its items ask for the techniques the medical OSCE checklist rates (paraphrasing, active listening, offering support), and all three end in a global rating. This is the common ground the transfer we measure can draw on.

\section{Method}
\label{sec:method}

Terms used throughout: each corpus is one \emph{domain}; a \emph{fold} holds one domain out and trains on the other two, so leave-one-domain-out (LODO) has three folds. A \emph{battery} is a set of construct questions (e.g. verbalizing, paraphrasing, structuring), a \emph{judge} is one LLM run with one battery (e.g.\ gemma3-12B \citep{gemma3} answering the eight BGR+OSCE questions is one judge), and an \emph{ensemble} is a set of judges whose score vectors are concatenated. A \emph{configuration} is one feature set (including the judge membership) with one preprocessing, projection and regressor setting. The \emph{primary system} is one fixed configuration on the \emph{compact stack}: the construct scores of four judges plus the compact nonverbal block and its dyadic deltas; \emph{language-only} keeps the same preprocessing, projection and regressor settings on the four judges alone. Two \emph{training settings} are compared on the same sessions: \emph{cross-domain}, trained on the other two corpora, and \emph{in-domain}, plain shuffled 5-fold cross-validation inside the held-out corpus (stratification impossible under data scarcity).

\subsection{Feature families}
\label{sec:features}
All features are extracted with DISCOVER \citep{hallmen2025discover}: transcription and speaker diarization (for \textsc{Vacc} see Section~\ref{sec:data}), facial expression, gaze and smiling from the video, and acoustic and prosodic descriptors from the audio. All features are session-level scalars over the assessed role and, for dyadic features, the interlocutor (Table~\ref{tab:features}).
\begin{table*}[t]
\small\centering\setlength{\tabcolsep}{5pt}
\begin{tabular}{@{}>{\raggedright\arraybackslash}p{2.6cm}>{\raggedright\arraybackslash}p{8.3cm}r>{\raggedright\arraybackslash}p{3.3cm}@{}}
\toprule Block & What & dims & role \\ \midrule
LLM judges & construct scores of four judges (Section~\ref{sec:judges}) & 26 & primary (language-only, stack) \\
Compact (\texttt{cmp}) & pitch mean/std, loudness, sentiment, gaze/smile rates, happy/neutral fractions, words/s, speech-emotion V/A/D \citep{wagner2023dawn}, facial V/A & 14 & primary (stack) \\
Deltas (\texttt{dlt}) & assessed minus interlocutor, same 14 & 14 & primary (stack) \\
Nonverbal-dyadic & deltas, mutual gaze/smile (intersections of both speakers' intervals), turn-taking (latencies, interruption/overlap, share, turns/min, lengths), windowed prosodic synchrony & 39 & exploratory (alternative stack) \\
Wide functionals & eGeMAPS \citep{eyben2016gemaps} LLDs (88), OpenFace 2.0 \citep{baltrusaitis2018openface} AUs (17), EmotiEffLib \citep{savchenko2023facial} V/A (2) $\times$ 5 functionals; expression fractions (8), sentiment \citep{barbieri2022xlmt}, gaze/smile & 563 & exploratory baseline \\
Shallow transcript & WhisperX \citep{bain2023whisperx} segments, words/min, word length, questions, speaking share, segment duration, unfinished segments & 7 & exploratory baseline \\
Sentence embeddings & duration-weighted means \citep{reimers2019sbert}, MiniLM-L12-v2 & 384 & exploratory baseline \\
\bottomrule
\end{tabular}
\caption{Feature blocks. Stack = compact stack (judges, compact block and deltas, 54 columns); V/A(/D) = valence/arousal(/dominance).}
\label{tab:features}
\end{table*}
Transcripts, and so sentiment, words per second and turn timing, derive from the audio, as the acoustic features do: the only modality-clean split is audio versus video. The compact and nonverbal-dyadic blocks thus carry a few transcript-derived columns; Appendix~\ref{app:matchedn} reports a strictly audio-visual control, the nonverbal-dyadic block with its transcript-derived columns removed.

\subsection{Instrument-anchored LLM judges}
\label{sec:judges}
The core feature family works in two stages (Figure~\ref{fig:pipeline}). A small local LLM reads the \emph{entire} two-speaker transcript, time-ordered and speaker-labeled with domain-neutral tags (\texttt{B:} assessed, \texttt{G:} interlocutor, after German \emph{Bewertete(r)} and \emph{Gespr\"achspartner(in)}; no professions), and returns session-level construct ratings (1--5) as JSON; these scores are then features for the regressor of Section~\ref{sec:protocol}. Every judge receives the complete transcript (transcript cap 16k characters, never reached: longest transcript 15.2k).
\begin{figure}[t]
\centering
\includegraphics[width=\columnwidth]{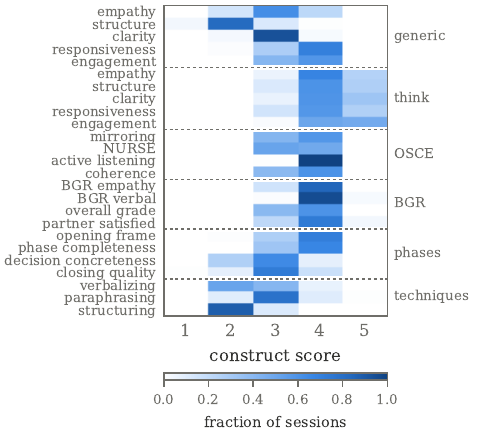}
\caption{Judge score distributions over the 195 sessions: three batteries in six groups (generic dialogue without and with thinking; BGR+OSCE as OSCE and BGR; phase+technique as phases and techniques); cross-family BGR+OSCE runs omitted (inventory in Table~\ref{tab:judges}).}
\label{fig:judgedist}
\end{figure}
Three batteries, all prompted in German, supply the constructs: \textbf{generic dialogue} (empathy, structure, clarity, responsiveness, engagement); \textbf{BGR+OSCE}, derived from the experts' instruments: BGR items (flipped so 5 = best), OSCE empathic techniques (e.g.\ NURSE \citep{back2005nurse}) from the medical communication curriculum \citep{zerbini2024komcua}, and a perspective-flip item (``how satisfied does \emph{G} leave?''); and \textbf{phase+technique}, from a counseling phase model and technique codebook \citep{hallmen-etal-2026-annotating,benien2003schwierige} (e.g.\ concreteness of the decision phase, paraphrasing). The primary ensemble has four judges \citep{qwen3,granite2026}: generic dialogue on gemma3-12B and on qwen3-14B in thinking mode, BGR+OSCE on gemma3-12B, and BGR+OSCE with anti-band wording on granite4.1-30B (an added instruction to use the full scale, Appendix~\ref{app:variants}). Sampling uses each model card's recommended parameters with pinned seeds (temperature $0.4$ for gemma3 and granite, $0.6$ for the thinking variant). The membership is chosen by nested selection (Section~\ref{sec:protocol}); Table~\ref{tab:judges} (Appendix~\ref{app:prompts}) inventories all eight full runs.

\paragraph{Constructs, not items.} Asking the judge for the constructs the experts rate is the design goal; the judge is zero-shot and never sees a rating. The batteries operationalize the instruments rather than replay them, and the mapping is deliberately not one-to-one (Appendix~\ref{app:mapping}). In prompt design we asked the instrument items directly, as rephrased constructs, and as a selected subset with rephrased level descriptions; the selection with rephrasing worked best, and on the 30-session screening subset the instruments' literal overall-item wordings alone reach $\rho = .26$--$.28$ against $.35$ for the battery's overall construct (screening-level evidence, Appendix~\ref{app:variants}). \textsc{Teach}, whose rating instrument the judges do not quote, is nevertheless the strongest fold.

\paragraph{Screening.} The regressor needs between-session variance, so before any full run a candidate model must pass a label-free gate on 30 pilot sessions: a parseable JSON object within three seeded attempts for $\ge 90\%$ of sessions and a mean per-construct standard deviation $\ge .35$. The gate excludes judges whose scores collapse to a constant, such as the scale ceiling; it reads outputs only, never targets (prompt wording was target-screened separately, Appendix~\ref{app:variants}). Admitted judges span 1--5 and concentrate at 3--4 like the expert ratings without collapsing (gate spread $0.37$--$0.80$); two BGR+OSCE constructs are near-constant (Figure~\ref{fig:judgedist}), and the variance threshold drops such a column where it is constant across the training sessions.

\subsection{Feature standardization}
\label{sec:zcal}
In every standardized feature set (Figure~\ref{fig:families}), each column is z-standardized, $z = (x - \mu)/\sigma$ with mean $\mu$ and standard deviation $\sigma$ taken from training sessions only and asymmetrically: each training corpus is standardized with its own $\mu$ and $\sigma$, the held-out corpus with the pooled raw $\mu$ and $\sigma$ of the two training corpora. No held-out feature statistics or ratings are used to fit a fold. The per-corpus statistics on the training side remove instrument and register offsets: a judge's ``3'' sits at a different point of the quality range in a clinical consultation than in a parent conversation. Corpus standardization precedes median imputation; in in-domain cross-validation all pipeline stages are fitted on the training four fifths, while the corpus standardization uses the corpus's own label-free feature statistics.

\subsection{Protocol}
\label{sec:protocol}

\paragraph{What is compared.} For each held-out domain both training settings predict exactly the same sessions ($n{=}57/53/85$ for \textsc{Ticks}/\textsc{Vacc}/\textsc{Teach}). The \emph{cross-domain} model is trained once on all sessions of the other two domains ($138/142/110$) and predicts every session of the held-out domain. The \emph{in-domain} reference is 5-fold cross-validation inside that domain: each session is predicted by a model trained on the other four fifths of its own corpus ($45$--$46/42$--$43/68$ sessions; test fifths of $11$--$12/10$--$11/17$), the out-of-fold predictions are concatenated, and the split is repeated over 20 random partitions ($\mu$ and $\sigma$ reported). Spearman's $\rho$ \citep{spearman1904proof} is computed per domain and averaged over domains: the targets are ordinal levels with ties, which suits a rank rather than a Pearson correlation, and pooling corpora with different rating distributions would reward predicting the corpus, not the session. The gap between the settings is estimated by a paired session bootstrap: $10^4$ resamples, stratified by domain, each session keeping its rating and both predictions, models fixed. On each resample the per-domain gap is the cross-domain $\rho$ minus the in-domain $\rho$ (mean over the 20 splits), then averaged over domains. The interval is conditional on these three corpora and excludes training variance. We call this, with the standardization of Section~\ref{sec:zcal}, the \emph{inductive protocol}. Two controls address the two obvious confounds. \emph{Training-size matched}: the cross-domain model is trained on $m = \mathrm{round}(0.8\,n)$ other-domain sessions, $n$ the held-out corpus size ($m = 46/42/68$; 200 draws), the training count of an in-domain fold model. \emph{Test matched}: the full cross-domain model and the in-domain fold model are scored on the same test fifth, for all five fifths of the same 20 splits; we report the gap averaged over fifths and splits, and the number of splits in which it is positive. The full-domain correlation and the mean of fifth-wise correlations are different estimands; we report both.

\paragraph{Configuration.} Feature set, projection and regressor are fixed by design: median imputation $\to$ variance threshold (drops constant columns) $\to$ standardization $\to$ PCA to 10 principal components (dimensions) $\to$ elastic net ($\alpha = 0.5$, $\ell_1$ ratio $0.5$, up to 5000 iterations) \citep{wold1987principal,zou2005elasticnet}, every stage fitted on training sessions only; the result is identical at every requested width from 1 to 50 (the effective width is capped by the column count; Appendix~\ref{app:matchedn}). The judge-ensemble membership is chosen by nested selection: in each outer fold, every candidate membership (Appendix~\ref{app:matchedn}, Membership selection; picks in Table~\ref{tab:nested}) is scored on the inner leave-one-training-domain-out split of the two training corpora, under the same training-statistics standardization, and the best inner score is kept; the held-out corpus never informs the choice. The picks are reported with the results (Section~\ref{sec:results}); we fix the same four judges for both systems for comparability. Imputation handles missing values with identifiable causes (Appendix~\ref{app:matchedn}). Code, prompts and out-of-fold predictions are linked in Appendix~\ref{app:repro}.

\section{Results}
\label{sec:results}

\input{primary_table}
\begin{figure}[t]
\centering
\includegraphics[width=\columnwidth]{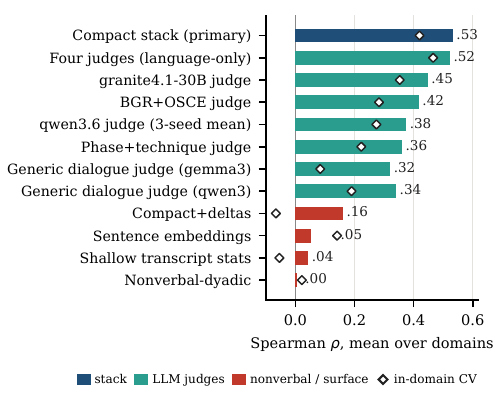}
\caption{Feature families at the fixed configuration under the inductive
protocol: cross-domain $\rho$, mean over held-out domains (bars), and the
20-split in-domain mean (diamonds); the judge-ensemble, stack, nonverbal-dyadic
and shallow rows use the standardization of Section~\ref{sec:zcal}, the
others raw columns. The qwen3.6 and phase+technique judges are not in the
primary system. Per-fold values are in Table~\ref{tab:main}
(Appendix~\ref{app:grids}).}
\label{fig:families}
\end{figure}

All numbers use the de-rounded \textsc{Teach} target (Appendix~\ref{app:equating}) and the repaired \textsc{Vacc} attribution (Section~\ref{sec:repair}).

\paragraph{Cross-domain training exceeds in-domain training for the compact stack.} Trained on the other two corpora, the primary system reaches $\rho = .533$ (folds $.485/.517/.598$, Table~\ref{tab:primary}); trained inside the held-out corpus and scored on the same sessions it reaches $.420$ (20 splits; per-domain $\sigma$ $.030$--$.054$). The paired bootstrap puts the gap at $+.113$, 95\% CI $[+.067, +.157]$, one-sided bootstrap $p \le 10^{-4}$ (no resample at or below zero), and the cross-domain model is ahead in every domain.

\paragraph{The advantage persists under training-size and test-grouping controls.} Subsampled to the in-domain training count, the cross-domain stack reaches $.506$ ($.486/.426/.607$), above the in-domain reference in every domain, a gap of $+.086$ at approximately equal training size; the \textsc{Vacc} value varies most across draws ($\sigma = .098$). Scored on identical test fifths, the gap is $+.065$, positive in 19 of 20 splits ($+.098/+.070/+.025$ per domain). The stack's advantage therefore persists under each control separately. The result does not depend on the per-corpus standardization step: without it (raw columns, keeping only the pipeline's standardization fitted on the training sessions) the stack reaches $.551$ and language-only $.521$, with gaps of $+.131$ $[+.089, +.173]$ and $+.055$ $[+.037, +.075]$ (Appendix~\ref{app:matchedn}); the protocol keeps the pre-specified training-statistics variant (Section~\ref{sec:zcal}).

\paragraph{Nested selection confirms the membership.} For the compact stack the inner selection picks the same four judges in every fold (Table~\ref{tab:nested}), so the primary system is one fixed membership and its nested and fixed results coincide ($.533$). For the language-only system the \textsc{Vacc} fold's inner selection adds two judges (strict nested $.495$ against $.523$ with the fixed four).

\paragraph{Language-only and in-domain training differ little on identical test fifths; the stack exceeds both.} The four judges alone transfer at $.523$ ($.485/.486/.597$) against $.466$ in-domain, a full-domain gap of $+.057$ $[+.038, +.077]$ and $+.042$ at matched training size. On identical test fifths, the mean gap is small ($+.006$, positive in 14 of 20 splits; $+.014/-.005/+.007$ per domain). The judges also fit better in-domain than the stack does, so their smaller cross-domain margin mostly reflects stronger in-domain fitting ($.466$ against $.420$) rather than weaker transfer ($.523$ against $.533$). The block increment, stack minus language-only on the same sessions, is $+.011$ $[-.019, +.040]$ (one-sided bootstrap $p = .24$), almost entirely in the \textsc{Vacc} fold ($.000/+.031/+.001$): the four judges carry the signal.

\paragraph{Language dominates the feature families.} In Figure~\ref{fig:families} every single judge transfers at $.32$--$.45$ and the four-judge ensemble at $.523$, while the compact nonverbal block with its deltas reaches $.161$ and the nonverbal-dyadic block, the shallow transcript statistics and the sentence embeddings stay at or below $.06$ at the fixed configuration; in-domain, no nonverbal or surface block exceeds $.15$.

\paragraph{The constructs matter more than the quoted level descriptions.} At the same model and transcript, asking the experts' constructs instead of generic dialogue qualities raises transfer from $.321$ to $.418$, a comparison that also varies construct count and prompt length; quoting the instruments' level descriptions, with the battery fixed, adds $+.004$ in a controlled comparison under the same protocol, within sampling-seed variation (Table~\ref{tab:anchor}, Appendix~\ref{app:mapping}). Judges from three model families ensemble to $.523$, above the strongest single judge, granite4.1-30B at $.448$; for the compact stack the inner selection picks the same four judges in every fold; for language-only the \textsc{Vacc} fold adds qwen3.6 and phi4 (Table~\ref{tab:nested}). Alone, the qwen3.6 judge transfers at $.375$ and the phase+technique judge, which was not a candidate, at $.361$.

\paragraph{Single-source transfer.} Trained on one corpus at a time (Table~\ref{tab:single}), every directed cell is positive, and the harder test pair \textsc{Teach}--medical transfer better ($.558$ stack, $.536$ language-only) than the medical pair ($.477$, $.487$), which suggests the result does not rest on the two medical corpora resembling each other, and a single corpus of 53--85 sessions already transfers at $.47$--$.63$.

\input{single_source_table}

\paragraph{Ranking, not scoring.} Predictions are rank-associated with the expert ratings ($\rho = .53$) but compress the scale. Mean absolute error (MAE) rewards mean-collapse; absolute-error and calibration results are in Appendix~
ef{app:errors}. The mean prediction rises monotonically with the rating from level 2 upward ($3.28$, $3.59$, $3.75$, $3.94$ for rounded true levels $2$--$5$; level 1 has two sessions), quadratic-weighted $\kappa$ is $.33$, and MAE is $.702$ against $.815$ for a constant at the pooled target mean (an oracle that uses the held-out targets). Like any regularized regressor, it shrinks predictions toward the center: their $\sigma$ is $0.44$ against $0.99$, and the level means span only $3.28$ to $3.94$ of the five-point range (Appendix~\ref{app:errors}). It should rank sessions and flag the lowest, not report instrument-scale levels without calibration.

\subsection{Recording Attribution}
\label{sec:repair}

\paragraph{What correct speaker attribution is worth.} \textsc{Vacc} lost its per-role audio (Section~\ref{sec:data}). Against its human speech-activity annotation, \textbf{16.1\%} of diarized segments carried the wrong speaker; restoring the attribution lifts the language-only \textsc{Vacc} fold from $.402$ to $.486$ and the language-only mean from $.493$ to $.523$, and the compact stack's \textsc{Vacc} fold from $.414$ to $.517$ and its mean from $.502$ to $.533$, since those sessions also train the other two folds (Appendix~\ref{app:equating}).

\FloatBarrier

\section{Discussion}
\label{sec:discussion}

\paragraph{How high is the ceiling?} An out-of-domain Spearman of $.53$ should be read against how well \emph{humans} agree, not against 1.0. Our targets are effectively single-rater; under a classical measurement-error reading \citep{spearman1904proof}, $\sqrt{\mathrm{ICC}}$ is a rough reliability reference rather than a strict ceiling for a Spearman correlation: $.73$ on \textsc{Teach} (single-rater ICC(2,1) $=.53$, from a parallel item) and $.88$--$.92$ on the medical corpora (double-rated subset ICCs $.77$--$.85$). On the medical corpora the model ($.485$/$.517$ on \textsc{Ticks}/\textsc{Vacc}) also agrees with the expert better than the other people in the room do (standardized patient $.33$/$.47$, student self-rating $.31$/$.29$; Appendix~\ref{app:constructs}).

\paragraph{Why does the transcript carry so much?} The expert instruments are largely language-defined, and the one construct the judges cannot see, BGR's nonverbal item, is not required to reach $.523$; whether richer nonverbal representations would add more is untested. Adding the compact nonverbal block gives $.533$ ($+.011$, interval including zero), so the judges carry the signal. Very poor nonverbal behavior would also show in how the conversation goes, which may be why the block adds little beyond the transcript.

\paragraph{Deployment view.} Learners on these programs ask for interpreted, construct-level statements rather than bare numbers \citep{bauermann2025acceptance,gietl2026foerderung}; a predictor that carries the expert's judgment into a program that trained none of it is the step before that. The ordering it needs is supported ($\rho = .53$, quadratic-weighted $\kappa = .33$, mean predictions monotone in the expert level from level 2 upward; Section~\ref{sec:results}), enough to flag which sessions to look at first; the construct scores suggest what to look at, their item-level validity is future work.

\section{Conclusion}
Across three German counseling corpora, an expert's overall impression of a session is predictable in a domain that contributed no training data: instrument-anchored LLM judges with a compact nonverbal block reach $\rho = .53$ against $.42$ in-domain, with most of the signal in the transcript.

The judges and the nonverbal feature models are small, open-weight and run on institutional hardware, so no recording leaves it, preempting privacy concerns. The judges read transcripts and the nonverbal block adds little, and the attribution repair motivates reliable per-speaker audio capture, for instance a lapel microphone per speaker (Section~\ref{sec:repair}). Therefore a less intrusive audio-only recording setup without cameras suffices. Programs with few rated sessions can start from other programs' data, which here trained the compact stack better than their own, and counseling settings beyond medical and teacher education whose instruments share these constructs are candidates. A rank-valid predictor shows which learners to look at first, and a low construct score suggests what to practice. Establishing whether those scores are valid item by item and help learners requires a long-term controlled study; calibration to levels is future work.

\section*{Limitations}
\textbf{Three domains.} The cross-domain estimate rests on three folds whose values vary ($.485$--$.598$ stack, $.485$--$.597$ language-only); the bootstrap interval is conditional on these corpora, and transfer to a fourth domain is untested. \textbf{Single-rater targets.} Each session carries one expert's rating, mostly at 3--4, and pairwise exact agreement between experts on overlap subsets is $.42$--$.47$, which limits attainable correlation (Section~\ref{sec:discussion}). \textbf{Setting.} All sessions are simulated, German-language role-plays from one university; real consultations, other languages and other rating cultures are untested. In-domain folds are not grouped by actor; actor-grouped in-domain values (\textsc{Vacc}, \textsc{Teach}; actor-grouped CV, not tabulated) move language-only from $.423$ to $.401$ (\textsc{Vacc}) and $.559$ to $.535$ (\textsc{Teach}), and the stack from $.380$ to $.205$ and $.533$ to $.501$. \textbf{Target coding.} The \textsc{Teach} target re-spreads rounded four-point answers under an equal-interval assumption (Appendix~\ref{app:equating}). \textbf{Modality.} The nonverbal increment ($+.01$) is not modality-clean, since the compact block carries transcript sentiment and speech rate (Section~\ref{sec:features}); a strictly audio-visual control of the nonverbal-dyadic block is in Appendix~\ref{app:matchedn}. \textbf{Ranking, not scoring.} The predictor orders sessions; reporting levels on the instrument's scale needs recalibration first (Appendix~\ref{app:errors}). \textbf{Wording and model family.} Within the primary ensemble, the cross-family judge (granite4.1-30B) also differs from the gemma3 BGR+OSCE judge in prompt wording (the anti-band instruction), so there the two effects are not separated (Appendix~\ref{app:mapping}).

\section*{Ethical Considerations}
All conversations are simulated role-plays between consenting adult students and trained actors or peers, recorded within the regular curricula of the students' degree programs; the underlying data collections have documented ethics approvals and consent procedures in the respective corpus publications \citep{zerbini2025multimeasure,schneider2025smiling,hallmen-etal-2026-annotating}. Automatic communication scoring is intended for formative feedback, not for high-stakes decisions; we report domain transfer precisely because deploying an in-domain-trained scorer on new populations without validation would be unsafe. No personally identifying data leaves the institution's infrastructure; all LLM inference runs on open-weight models on institutional hardware. We have not established that the scores are equally valid across speaker groups: the corpora are too small and their demographic records too incomplete to estimate subgroup performance, the judges' pretraining may carry register and accent priors, and the diarization error we measure in one corpus need not fall evenly across speakers. Differential validity is therefore untested, which is a reason to keep such scores formative and paired with human review.

\section*{Acknowledgments}
This work was partially funded by the KodiLL project (FBM2020, Stiftung Innovation in der Hochschullehre).

\bibliography{refs}

\clearpage
\appendix
\section{Data and Target Integrity}
\label{app:equating}
\paragraph{The attribution repair in detail.} \textsc{Vacc}, the corpus whose per-role streams were lost (Section~\ref{sec:data}), carries human speech-activity annotation for all 53 sessions. Re-attributing every diarized segment to the role whose annotated speech overlaps it most shows that \textbf{16.1\%} carried the wrong speaker. Restoring the attribution lifts the granite judge's strongest constructs markedly ($+.08$ on the overall grade, $+.11$ on partner satisfaction) and the language-only \textsc{Vacc} fold from $.402$ to $.486$, the language-only mean from $.493$ to $.523$ (compact stack: $.414$ to $.517$ and $.502$ to $.533$), since those sessions also train the other folds. That is what the microphone failure cost: a recording-setup decision priced in predictive validity rather than in diarization error rate. The repair propagates to every transcript-derived feature, turn-taking included, but not to the voice features, which have no per-role channel to come from; face and gaze were never affected. It errs in both directions: segmentation and overlap errors still come from the mixed signal, while human annotation attributes more cleanly than a second channel with bleed would.

\begin{table}[ht]
\centering\small
\begin{tabular}{lrr}
\toprule
Placement of the four-point era & $\rho$ & ties \\
\midrule
$\{1,3,4,5\}$ as stored & .44 & 33.2\% \\
$\{1,2,4,5\}$ midpoint-skipping & .47 & 30.8\% \\
$\{1,2.33,3.67,5\}$ equal-interval & \textbf{.53} & 21.0\% \\
$\{1,2,3,4\}$ shifted below & .58 & 30.4\% \\
equipercentile$^{\dagger}$ & .40 & 50.3\% \\
\midrule
within-era reference & .51 & --- \\
\bottomrule
\end{tabular}
\caption{Sensitivity of the \textsc{Teach} fold to how the four-point era
is placed on the five-point range, at one fixed reference configuration
(an earlier development configuration: three judges plus the compact block,
elastic net on 10 principal components). The
within-era reference is invariant to the placement, since every candidate
is monotone within an era; a pooled value below it indicates destroyed
cross-era ordering, above it manufactured separation. $^{\dagger}$The
equipercentile map degenerates at $n{=}24$, sending two four-point
categories to the same value, which is why it also depresses the
within-era reference (to $.44$) rather than only the pooled score.}
\label{tab:equating}
\end{table}

The second artifact is in the target itself. \textsc{Teach} switched its overall item from five to four points after the first 24 of its 85 sessions, and the four-point answers were stored as $\mathrm{round}(x \cdot 5/4)$, i.e.\ $\{1,2,3,4\} \mapsto \{1,3,4,5\}$. The map is injective, and it is a defensible equating rather than a slip: it aligns the four-point categories with the top four five-point ones, and is within two session pairs of what one obtains by collapsing the five-point era's near-empty bottom category ($\rho = .99998$; exactly one of those 24 sessions was ever rated 2, and none 1). What it also is, however, is the equating that maximizes ties. Four categories placed on an integer five-point grid must collide with categories the five-point era already occupies, and because both eras sit in one LODO fold the rank metric is computed \emph{across} them: $33.2\%$ of session pairs in the fold are tied in the target, capping the Spearman a strictly ordered prediction can reach on that fold at $.93$ (a predictor that reproduced the ties exactly could still reach $1$). It is not a level distortion, the two eras' means under this coding being $3.79$ and $3.90$, but a loss of resolution.

No session was rated on both versions of the target item, so no common-item design is available and no coding can be validated directly. Two facts let us choose anyway. First, any coding is monotone \emph{within} an era, so within-era rank performance is invariant to the choice and serves as a coding-free reference: a pooled coding that scores below it is destroying cross-era ordering, and one that scores above it is manufacturing separation the model cannot demonstrate within either era. Table~\ref{tab:equating} scores the candidates against that reference. The equal-interval coding is the only one that lands within $.02$ of it; the coding that shifts the four-point era below the five-point one scores highest of all and is thereby identified as over-correction, not repair. Second, the four expert raters also scored the 24 five-point sessions on a separate four-point instrument (Appendix~\ref{app:raters}); their mean, $2.98$, sits essentially where the four-point era itself sits ($2.93$) and where a linear rescaling of the five-point ratings puts it ($3.09$). An instrument that was natively four-point and never rescaled thus places the two eras at the same level independently, which is what an over-correcting coding has to deny.

We therefore invert the rounding and re-spread the four-point answers over the five-point range, leaving the 24 five-point sessions untouched. This raises the fold's attainable Spearman from $.93$ to $.97$, mechanical bookkeeping since any interpolating coding breaks ties. At that configuration, de-rounding against the stored coding moves the three-fold mean by about $+.03$; the residual choice among reasonable interpolating codings moves it by under $.01$. The same inversion exposed one session whose rescale had been missed and later ``corrected'' to the wrong neighbor; we restore it from the surviving raw four-point sheet.

\paragraph{Rater Designs and Agreement.}
\label{app:raters}
In \textsc{Ticks} one expert rated all sessions and a second independently rated 15\% of them (ICC $= .80$) \citep{zerbini2025multimeasure}. In \textsc{Vacc} four experts (experienced OSCE assessors with a long-shared, partly consensual rating practice, a pool that also includes the \textsc{Ticks} rater) split the analyzed sessions (13--14 of the 53 each); reliability on double-rated subsets is ICC $= .77$--$.85$ \citep{schneider2025smiling}, yet on the six-session calibration overlap (five of them rated by all four, one by two) pairwise exact agreement on the overall item is $.42$, within-one $.94$, and Krippendorff's ordinal $\alpha = .24$ over 31 rater pairs, deflated by the overlap's restricted variance. In \textsc{Teach} the target comes from a pool of four school counseling experts (one rating per session); the same four each rated a 24-session subset on an overall-helpfulness item (100\% overlap there): mean pairwise $\rho = .56$, Krippendorff's $\alpha = .52$, ICC(2,1) $= .53$, average-raters ICC(2,$k$) $= .82$, pairwise exact agreement $.47$, within-one $.97$. The four-rater mean on that subset correlates with the per-session target at $\rho = .57$, about $.86$ once corrected for the attenuation implied by those reliabilities, i.e.\ the same construct read through two noisy instruments.

\section{Judge Prompts and Variants}
\label{app:prompts}
\input{judge_table}

All judges receive the merged, time-ordered two-speaker transcript with domain-neutral labels (\texttt{B:} assessed person, \texttt{G:} interlocutor), capped at 16k characters (never reached), and must answer with a single JSON object; parse failures trigger up to two retries with incremented seeds. Sampling uses each model card's recommended parameters (temperature $0.4$ for gemma3, granite and phi4, $0.6$ for the thinking variant, $0.7$ for qwen3.6) with pinned seeds. Prompts are German (the corpus language); the BGR+OSCE and phase+technique judges inline abridged level descriptions derived from the expert instruments. The generic dialogue judge (gemma3-12B, and qwen3-14B in thinking mode):

\begin{lstlisting}
Du bewertest die Gesprächsführung der Person "B" anhand des folgenden Gesprächstranskripts. "B" ist die zu bewertende Person, "G" ist ihr Gesprächspartner. Bewerte NUR Person B auf einer Skala von 1 (sehr schlecht) bis 5 (sehr gut) in diesen Dimensionen:
- empathie: einfühlsam, geht auf Gefühle von G ein
- struktur: klar gegliedert, roter Faden
- klarheit: verständlich, präzise formuliert
- eingehen: reagiert auf das von G Gesagte, greift es auf, stellt Fragen
- engagement: aktiv, zugewandt, engagiert
Sei streng und differenziert [...] Antworte AUSSCHLIESSLICH mit einem JSON-Objekt.
\end{lstlisting}

\noindent\emph{English translation:} ``You rate the conversational conduct of person `B' based on the following transcript. `B' is the person being assessed, `G' is their interlocutor. Rate ONLY person B on a scale from 1 (very poor) to 5 (very good) on these dimensions: empathy (sensitive, responds to G's feelings); structure (clearly organized, common thread); clarity (comprehensible, precisely phrased); responsiveness (reacts to what G said, picks it up, asks questions); engagement (active, attentive, committed). Be strict and discriminating [\dots] Answer EXCLUSIVELY with a JSON object.''

The BGR+OSCE judge (gemma3-12B) quotes the BGR items and the OSCE checklist's level descriptions:

\begin{lstlisting}
Du bewertest die Gesprächsführung der Person "B" anhand des folgenden Gesprächstranskripts. "B" ist die zu bewertende Person, "G" ist ihr Gesprächspartner bzw. ihre Gesprächspartnerin. Bewerte NUR Person B nach diesen Kriterien aus Expertenbögen für Gesprächsführung (alle Skalen: 5 = am besten):
- spiegeln: Paraphrasieren (Gesagtes in eigenen Worten wiedergeben) / Verbalisieren (vermutete oder geäußerte Emotionen benennen). 1 = spiegelt nie; 2 = an wenig treffender Stelle oder sprachlich hölzern; 3 = an treffender Stelle, aber wenig authentisch/elaboriert; 4 = an treffender Stelle, elaboriert und authentisch; 5 = zusätzlich als fragendes Angebot, sodass G sich dazu äußern kann.
- nurse: benennt Emotionen von G, zeigt Verständnis für die Lage/Sicht von G, respektiert Gefühle/Sicht von G, bietet Unterstützung an, exploriert bei Unklarheit. 1 = nichts davon; 2 = ein Aspekt an wenig treffender Stelle oder hölzern; 3 = ein Aspekt, wenig authentisch/elaboriert; 4 = ein Aspekt an treffender Stelle, elaboriert und authentisch; 5 = mehrere Aspekte an treffenden Stellen, elaboriert und authentisch.
- aktives_zuhoeren: offene Fragen, ausreden lassen, Hörsignale, zur Weiterrede ermutigen, Nachfragen, Gehörtes zusammenfassen. 1 = kein aktives Zuhören; 2 = selten oder an unpassender Stelle; 3 = mehrfach, aber wenig Variation; 4 = häufig und vielfältig, aber noch ausbaufähig; 5 = durchgängig angemessener, vielfältiger Einsatz.
- kohaerenz: Stimmigkeit und roter Faden der Gesprächsführung. 1 = keine erkennbare Struktur, G muss die Führung übernehmen; 3 = schablonenhaft, minimal flexibel oder inkonsistente Führung; 5 = sehr gute Struktur, Kontrolle über das Gespräch, transparente Überleitungen, flexibel.
- bgr_empathie: 5 = geht durchgehend verständnisvoll auf Hinweise und Bedürfnisse von G ein bzw. reagiert angemessen; 1 = geht nicht auf offensichtliche Hinweise/Bedürfnisse ein oder reagiert unangemessen.
- bgr_verbal: 5 = kommuniziert so, dass G ihn leicht versteht (Wortwahl, angemessene Ausdrucksweise); 1 = erschwert das Verstehen.
- gesamtnote: Gesamteindruck des Gesprächs als invertierte Schulnote, 5 = sehr gut, 1 = mangelhaft.
- partner_zufrieden: Wie zufrieden verlässt G vermutlich das Gespräch? 5 = sehr zufrieden, 1 = sehr unzufrieden.
Sei streng und differenziert: 3 bedeutet durchschnittlich, 5 nur ohne erkennbare Schwächen. Nutze die gesamte Skala. [Transkript] Antworte AUSSCHLIESSLICH mit einem JSON-Objekt, keine Erklärung.
\end{lstlisting}

\noindent\emph{English translation (constructs):} mirroring (paraphrasing / verbalizing emotions; 1 = never mirrors \dots\ 5 = additionally offered as a question so G can respond); NURSE (names G's emotions, shows understanding, respects G's view, offers support, explores when unclear; with level descriptions 1--5); active listening (open questions, letting finish, listener signals, encouraging, summarizing; 1 = none \dots\ 5 = consistently varied); coherence (common thread and control of the conversation; 1 = no discernible structure, G must take over \dots\ 5 = transparent transitions, flexible); BGR empathy (5 = consistently responds understandingly to G's cues and needs; 1 = ignores obvious cues or reacts inappropriately); BGR verbal expression (5 = communicates so G understands easily; 1 = impedes understanding); overall as inverted school grade (5 = very good); partner satisfaction (how satisfied does G presumably leave, 1--5).

The phase+technique judge (gemma3-12B) quotes the phase-model and technique definitions of the annotation guideline:

\begin{lstlisting}
Du bewertest die Gesprächsführung der Person "B" anhand des folgenden Gesprächstranskripts eines Beratungsgesprächs. [...] Grundlage: Gesprächsphasenmodell (Benien) und Gesprächstechniken (Verbalisieren, Paraphrasieren, Strukturieren). Phasen dürfen in der Reihenfolge variieren und mehrfach vorkommen; bewerte Vorhandensein und Qualität, nicht starre Abfolge. Alle Skalen: 1 = fehlt/sehr schlecht, 5 = vorbildlich.
- anfang_rahmen: Anfangsphase — Begrüßung, ggf. Smalltalk, zeitlicher Rahmen (Zeitangabe) und inhaltlicher Rahmen (Anliegen/Agenda) werden geklärt. 5 = Kontakt hergestellt UND Zeit- und Inhaltsrahmen explizit geklärt.
- phasen_vollstaendig: Sind die der Situation angemessenen Phasen erkennbar (Anfang, Information, Argumentation/Lösungssuche, Beschluss, Abschluss)? 5 = alle klar erkennbar und sinnvoll gefüllt; 1 = Gespräch bleibt in einer Phase stecken oder Phasen fehlen.
- beschluss_konkret: Beschlussphase — werden Vorschläge in konkrete Handlungsschritte übersetzt, ggf. (SMARTe) Ziele oder Vereinbarungen? 5 = konkrete, gemeinsam getragene nächste Schritte; 1 = nichts vereinbart.
- abschluss_qualitaet: Abschlussphase — wertschätzende Reflexion, ggf. Terminvereinbarung, Verabschiedung. 5 = würdigender, runder Abschluss; 1 = abruptes Ende.
- verbalisieren: B fokussiert wahrgenommene Gefühle von G — kurze Aufmerksamkeitsreaktionen, direktes Benennen eines Gefühls, klärende Fragen bei vagen Andeutungen, weiterführende emotionale offene Fragen. 5 = mehrfach, treffend, vertiefend.
- paraphrasieren: B gibt die sachliche Kernaussage von G in eigenen Worten wieder (ggf. verknappt). 5 = mehrfach und treffend an wichtigen Stellen.
- strukturieren: B wechselt auf die metakommunikative Ebene, um Ablauf zu steuern und transparent zu machen — Agenda, Überleitungen, Zusammenfassen des Gesagten. 5 = durchgängig klare, transparente Steuerung.
Sei streng und differenziert [...] Antworte AUSSCHLIESSLICH mit einem JSON-Objekt, keine Erklärung.
\end{lstlisting}

\noindent\emph{English translation (constructs):} opening frame (greeting, small talk if appropriate, explicit time frame and content frame/agenda); phase completeness (are the situation-appropriate phases discernible: opening, information, argumentation/solution search, decision, closing; 1 = conversation stuck in one phase); decision concreteness (proposals turned into concrete, jointly supported next steps or (SMART) agreements); closing quality (appreciative reflection, follow-up appointment if appropriate, farewell; 1 = abrupt ending); verbalizing (focuses G's perceived feelings: attention signals, naming a feeling, clarifying questions on vague hints, deepening open emotional questions); paraphrasing (restates G's factual core message in own words); structuring (switches to the metacommunicative level to steer and make the process transparent: agenda, transitions, summarizing). Phases may vary in order and recur; presence and quality are rated, not rigid sequence.

\paragraph{Generation, budget and screening details.} Generation uses model-card sampling parameters with pinned seeds and seed-bumped retries; all 195 sessions parse within three attempts for every judge. The prompt caps the transcript at 16k characters, which no session reaches: the longest transcript is 15.2k characters and the longest prompt built from it 18k. Nor did any run approach the generation budget: the think-mode judge, the only configuration where prompt and budget could collide, spends 455--780 generated tokens on its seven longest sessions, whose prompts measure 4186--4469 tokens, well below the 8192-token context we pin. The screening gate is on \emph{between}-session variance, whether a judge separates conversations at all; \emph{within}-session variance, whether it repeats its own score, is measured separately (Section~\ref{sec:judges} and Appendix~\ref{app:mapping}). The gate excludes ceiling collapse rather than certifying quality; survivors clear both, at $100\%$ parse and $0.37$--$0.80$ spread. Four negative results: gemma4-12B \citep{gemma4} and granite4.1-8B collapse to the scale ceiling despite the level descriptions, and reasoning modes repeatedly break the JSON contract (qwen3.5-27B mostly unparseable; qwen3.6-27B \citep{qwen3.6} \emph{in think mode} parses only 63\%).

\paragraph{Judge-Variant Screening.}
\label{app:variants}
On a 30-session screening subset (10 per domain) we compared prompt and model variants of the BGR+OSCE judge against its cached run, using per-domain standardized scores against the expert target. (i)~An added instruction to use the full scale and differentiate between dimensions improved the strongest constructs markedly (coherence $.35 \to .53$, mirroring $.30 \to .48$, overall grade $.35 \to .48$). (ii)~Running the BGR+OSCE prompt in a reasoning model (qwen3-14B, thinking mode) was unstable: single constructs improved while others turned negative. (iii)~Asking the two literal overall-item wordings of the instruments alone ($\rho = .26$--$.28$) did not outperform the battery's overall-grade construct ($.35$); a bare target question is weaker than a construct battery.

Variant (i) was then run on all 195 sessions, and the screening estimate does not survive at full scale: per construct the gains shrink and change sign (coherence $.35 \to .39$, overall grade $.34 \to .38$, BGR empathy $.13 \to .21$, but mirroring $.33 \to .29$ and NURSE $.25 \to .18$). The inner selection of Appendix~\ref{app:matchedn} does not pick the anti-band gemma3 judge in any fold. This is the clearest single illustration of why the screening subset cannot be used to set a headline: an improvement of up to $+.18$ per construct on 30 sessions does not survive at full scale.

\section{What the Judges Measure}
\label{app:mapping}
The judge batteries are derived from the expert instruments, not transcribed from them, and the mapping is many-to-one in places. \textbf{BGR} contributes three constructs directly (\emph{bgr\_empathie}, \emph{bgr\_verbal}, and \emph{gesamtnote} for the overall impression, all flipped so 5 = best); its structuring aspect is carried by \emph{kohaerenz}, which is worded from the OSCE checklist's level descriptions, and its nonverbal item is dropped, since a transcript-reading judge cannot observe it, which is precisely the modality gap Section~\ref{sec:results} measures. The \textbf{OSCE checklist} of empathic techniques contributes \emph{spiegeln}, \emph{nurse}, \emph{aktives\_zuhoeren} and \emph{kohaerenz}, each with the checklist's own level descriptions abridged to prompt length. The \textbf{phase model} \citep{benien2003schwierige} contributes four constructs rather than one per phase: its five phases enter through \emph{phasen\_vollstaendig} (are the situation-appropriate phases present and filled), while the opening, decision and closing phases additionally get a quality item each, since those are the phases the codebook describes most concretely. One construct is added that no instrument contains, \emph{partner\_zufrieden} (``how satisfied does G presumably leave?''), as a perspective flip on the same session.

Three arguments defend this over literal replay. \emph{Empirically}, the instruments' own overall-item wordings, asked alone, correlate at $\rho = .26$--$.28$ on the screening subset against $.35$ for the battery's overall construct (screening-level evidence, Appendix~\ref{app:variants}); and among the battery's constructs it is the behavior-level ones (phase completeness, coherence) that transfer across domains, while the most directly quoted scale items (BGR empathy, BGR verbal) are the unstable ones (Table~\ref{tab:constructs}). The instruments thus help at the level of observable behavior, not of scale labels. \emph{Structurally}, a rating instrument is written for a trained human who watches the whole interaction after a calibration session and may consult the manual; a zero-shot judge has one self-contained prompt, no visual channel and no calibration, so the level descriptions must be inlined and the visual items dropped. \emph{Methodologically}, we do not claim measurement equivalence with the instruments: the judge scores are features whose validity is argued predictively, against held-out expert ratings in domains that contributed no training data. An item-level equivalence claim would additionally require a within-domain agreement study against the instrument's own trained raters, which our rating designs (Appendix~\ref{app:raters}) do not support.

\paragraph{Model family and prompt wording.} The BGR+OSCE battery has one original-wording run (gemma3-12B) and four anti-band runs (gemma3-12B, qwen3.6-27B, granite4.1-30B, phi4-14B; Table~\ref{tab:judges}). Within the primary ensemble the gemma3 BGR+OSCE judge uses the original wording and granite4.1-30B the anti-band wording, so there a model-family effect and a prompt-wording effect cannot be told apart. Outside the ensemble the anti-band runs separate them: alone, the anti-band gemma3 run transfers at $.421$ ($.36/.43/.47$), against $.448$ for granite4.1-30B at the same wording and $.418$ for the original-wording gemma3 run on the same model. The controlled comparison below holds the model fixed and varies only the level descriptions.

\paragraph{What quoting the level descriptions is worth, controlled.} The $.321 \to .418$ comparison in Section~\ref{sec:results} varies the construct count (5 vs 8), the constructs themselves, the prompt length and the presence of an overall-grade item, so it cannot attribute the gain; Table~\ref{tab:anchor} separates generic versus bare expert constructs ($.321 \to .414$) from adding the level descriptions ($.414 \to .418$). A third prompt isolates the level descriptions: same eight construct keys in the same order, same scale direction, same JSON contract, same strictness sentence, same model, transcripts and seeds, with each instrument-derived level description replaced by a one-line gloss ($2653 \to 919$ prompt characters).

\begin{table}[ht]\centering\small
\begin{tabular}{lcc}
\toprule
Prompt & LODO $\rho$ & folds \\
\midrule
generic, 5 constructs & .321 & .34/.32/.30 \\
experts' 8 constructs, bare & .414 & .45/.36/.43 \\
\;\;$+$ quoted level descriptions & .418 & .40/.43/.42 \\
\bottomrule
\end{tabular}
\caption{Level-description control. All rows share model, transcripts and
seeds; only the last two rows share construct keys and differ solely in the
instrument-derived level descriptions.}
\label{tab:anchor}
\end{table}

\noindent Selecting the experts' constructs is worth $+.093$; quoting their level descriptions on top is worth $+.004$, well inside the seed noise of a single run. The instruments matter as a source of \emph{what to ask}, not as text to reproduce, which is consistent with literal item wordings transferring worse still (Appendix~\ref{app:variants}). Removing roughly two thirds of the prompt changes mean transfer by $-.004$.

\paragraph{Constructs against the target.}
\label{app:constructs}
\input{construct_table}

\paragraph{Construct-level transfer.} Table~\ref{tab:constructs} correlates each judge construct with the expert target per domain. The most domain-consistent constructs are \emph{phase completeness} (mean $\rho = .39$, weakest domain $.33$) and \emph{coherence} ($.35$/$.33$): experts across professions reward a conversation that moves through its phases under the professional's control. Directly quoted rating-scale items (BGR empathy, BGR verbal) are unstable across domains, while \emph{technique} constructs with behavioral level descriptions transfer; the instruments help at the level of observable behaviors, not scale labels. None of the constructs in Table~\ref{tab:constructs} has a cross-domain mean above $\rho = .39$, yet the primary language-only ensemble reaches $.523$: the judges form a battery of individually modest, partially decorrelated measurements that composite into a reliable score: the classical construction of a reliable instrument from unreliable items, not one single good question.

\paragraph{The judge as another rater in the room.} Read as a rater, the BGR+OSCE judge's overall grade agrees with the expert at $\rho = .32$/$.29$/$.41$ (\textsc{Ticks}/\textsc{Vacc}/\textsc{Teach}); the standardized patient reaches $.33$/$.47$/---, the student's self-rating $.31$/$.29$/---, and the calibrated four-expert panel $.57$ (\textsc{Teach}). A single zero-shot judge scores like the non-expert humans present in the conversation. On the medical corpora the trained model ($.485$/$.517$) surpasses the patient and self-ratings by combining many such judgments; on \textsc{Teach} it reaches $.598$, which is not directly comparable with the panel's $.57$ (a different item, on a 24-session subset).

\paragraph{How high is the ceiling, and what would the top of it mean?} Section~\ref{sec:discussion} reads $.53$ against $\sqrt{\mathrm{ICC}}$, which treats expert disagreement as measurement error attenuating an otherwise recoverable true score. That reading is standard but contested in two ways, and both matter for how the number should be reported. Empirically, annotator agreement has been shown not to cap system performance: models exceed it where the annotation task is harder for humans than the prediction task is for a model \citep{boguslav2017interannotator}. Conceptually, disagreement between competent experts on a subjective construct is better read as genuine variation than as noise around one true score \citep{plank2022problem,uma2021learning}, in which case there is no single ceiling to approach. Both readings cut the same way here. A model that agreed perfectly with one expert would be reproducing that expert's idiosyncrasy rather than the construct the instrument names, so we report $\sqrt{\mathrm{ICC}}$ as the scale our number should be read on, not as a target we are falling short of. It is also why the calibrated four-expert panel ($\rho = .57$ on \textsc{Teach}) is the more informative comparison than any single rater.

\paragraph{Judges are reproducible.} Judge variance decomposes on two axes that are easy to conflate. The screening gate of Section~\ref{sec:judges} asks whether a construct varies \emph{between} sessions, which is what makes it able to discriminate conversations; seed stability asks whether it repeats \emph{within} a session, which is measurement error. A useful judge maximizes the first and minimizes the second, and the two are the numerator and the denominator of a reliability ratio rather than competing goals. Across seeds the judges reproduce 80--98\% of their scores, except qwen3.6-27B at its higher sampling temperature (44--56\%), which therefore enters the candidate memberships as its three-seed mean (Appendix~\ref{app:matchedn}). Re-running the BGR+OSCE judge with two fresh sampling seeds on 30 sessions reproduces $91$--$95\%$ of construct scores exactly (mean pairwise $\rho = .75$--$.83$ per non-constant construct). Its disagreement is confined to adjacent scale points, so at the resolution the expert target carries it behaves as a near-deterministic feature extractor. We quote exact agreement rather than within-one because the bands are compressed: $99$--$100\%$ of all scores already lie within one point of the modal level, so a within-one reproducibility figure would mostly restate that compression.

\paragraph{Do speaker labels matter at all?} Feeding the merged transcript \emph{without} speaker tags and asking the judge to identify the counselor itself gains a little on the two medical corpora ($+.02$/$+.03$ mean construct correlation, since the doctor's register is identifiable from content) but costs $-.19$ in \textsc{Teach}, where two lay adults discussing a child are not separable by register.

\section{Protocol and Robustness Controls}
\label{app:matchedn}
Unless varied below, everything here uses the fixed configuration and the training-statistics standardization of Section~\ref{sec:zcal}. Imputation handles missing values with identifiable causes, not random gaps: absent per-role streams, one corpus's diarized audio without per-role acoustics, and one session whose face mask silently corrupted every facial feature while detector confidence stayed high.

\paragraph{Membership selection.} The candidate memberships are sets of judge runs of Table~\ref{tab:judges}. With \emph{base three} = \{generic dialogue on gemma3-12B, generic dialogue on qwen3-14B in thinking mode, BGR+OSCE on gemma3-12B\}, the candidates are: base three; base three $+$ \{anti-band gemma3-12B\}; base three $+$ \{granite4.1-30B\} (the primary membership); base three $+$ \{granite4.1-30B, qwen3.6-27B\}; and base three $+$ \{granite4.1-30B, qwen3.6-27B, phi4-14B\}, qwen3.6 entering as its three-seed mean. Language-only uses the judges alone. For the compact stack each membership enters with the compact block and its deltas; one candidate, base three $+$ \{granite4.1-30B, qwen3.6-27B\}, additionally includes 22 prosody functionals (11 F0, voice-quality, loudness and voicing functionals and their assessed-minus-interlocutor deltas); it is never picked. The nonverbal-dyadic stack pairs each membership, the prosody candidate included, with the nonverbal-dyadic block instead of the compact block. The inner rule: in each outer fold every candidate is fitted on one training domain and scored on the other, with the validation domain standardized by the inner-training domain's statistics, both directions averaged; the best inner score is applied once to the held-out domain, which never informs the choice (Table~\ref{tab:nested}). The compact stack picks the primary membership in all three folds, so its nested and fixed results coincide ($.533$). Language-only picks it in the \textsc{Ticks} and \textsc{Teach} folds and the largest membership in the \textsc{Vacc} fold ($.403$ there, $.495$ mean). The nonverbal-dyadic stack picks a different membership in each fold ($.488$).

\input{nested_table}

\paragraph{Standardization source.} The result does not depend on how the held-out corpus is standardized. Without the per-corpus standardization (raw columns, keeping only the pipeline's standardization fitted on the training sessions) the compact stack reaches $.551$ and language-only $.521$, and the paired transfer gap over in-domain training is $+.131$ $[+.089, +.173]$ and $+.055$ $[+.037, +.075]$. Standardizing the held-out corpus with a small unlabeled batch of its own sessions ($k = 10$ or $20$, the deployment case of a program with a cohort) gives $.541$/$.540$ for the stack and $.525$/$.524$ language-only. These target-batch variants are not strictly inductive; the protocol keeps the pre-specified training-statistics variant of Section~\ref{sec:zcal} ($.533$ and $.523$).

\paragraph{Training-size matched per domain.} Table~\ref{tab:matchedn} gives the per-domain values behind Table~\ref{tab:primary}: the cross-domain training set is subsampled to $m = \mathrm{round}(0.8\,n)$ other-domain sessions, $n$ the held-out domain's size, 200 draws, and compared with the 20-split in-domain reference at the same configuration. \textsc{Vacc} is the fold where subsampling costs most (stack $.517 \to .426$, $\sigma = .098$ across draws) and the only one where the size-matched model does not beat in-domain training in nearly every draw ($76\%$ of draws for the stack, $82\%$ language-only).

\input{matchedn_table}

\paragraph{Single-source transfer.} Table~\ref{tab:single} (Section~\ref{sec:results}) trains on one corpus and tests on another. The \textsc{Teach}--medical cells average $.558$ for the stack and $.536$ language-only, against $.477$ and $.487$ for the medical pair, so the pooled result is not carried by the two BGR-rated medical corpora resembling each other. The medical pair is the harder test pair: from the same 57 and 53 training sessions, \textsc{Ticks} and \textsc{Vacc} transfer to \textsc{Teach} at $.61$--$.63$ (stack).

\paragraph{A strictly audio-visual block.} The nonverbal-dyadic block times turn-taking from the transcripts and its deltas include speech rate and sentiment, so it is dyadic rather than non-linguistic. Removing every transcript-derived column leaves 26 of its 39 features, all audio-visual. Alone they reach $\rho = .182$ against $.004$ for the full block at the fixed configuration, so the transcript-derived columns were diluting the block rather than carrying it. Paired with the four judges, the strict block gives $.517$ and the full block $.519$, both at or below language-only ($.523$); neither version adds to the judges, and the modality attribution does not depend on the transcript-derived columns.

\paragraph{Projection width.}
\label{app:selectors}
The compact stack returns a mean $\rho$ of $.533$ and language-only $.523$ at every requested PCA width from 1 to 50 (the effective width is capped by the column count).

\paragraph{Per-family values.}
\label{app:grids}
Table~\ref{tab:main} gives the per-fold values behind Figure~\ref{fig:families}.

\input{main_table}

\paragraph{In-domain reference.} The in-domain reference is averaged over 20 random 5-fold splits because one split is unstable at $53$--$85$ sessions: the per-domain $\sigma$ across splits is $.030$--$.054$ for the stack and $.024$--$.031$ language-only (Table~\ref{tab:primary}), so any in-domain number at this $n$ should be reported with its spread. Neither a single split nor best-over-grid in-domain cells are used as the reference.

\FloatBarrier

\section{Error Analysis}
\label{app:errors}
Out-of-fold cross-domain predictions of the primary system are strongly compressed toward the scale center: the mean prediction rises monotonically with the expert rating from level 2 upward ($3.28$, $3.59$, $3.75$, $3.94$ for rounded true $2$ to $5$; $32$, $41$, $73$, $47$ sessions; level 1 has two sessions, $\mu = 3.52$), but spans only $0.7$ of the five-point range. Mean predictions increase across rating levels while individual ranking errors remain, and the level is not recovered, which is the regime in which rank metrics are informative and MAE is not: predicting the pooled mean everywhere would cost little absolute error and all rank information.

\input{calib_table}

\noindent MAE beats the mean-only baseline (a constant at the evaluated sessions' own target mean, an oracle that uses held-out targets) everywhere, but only by $.07$--$.17$, and quadratic-weighted $\kappa$ is $.33$ pooled, well below the rank correlation. The predictions' standard deviation is $0.44$ against $0.99$ for the targets, so the model keeps a rank association while compressing the scale by more than half. Any deployment reporting numbers rather than rankings would need to recalibrate first; the language-only ensemble behaves the same way (MAE $.707$, QWK $.307$, sd $0.47$).

The rank error is \emph{not} concentrated at the extremes: normalized within each domain it is $0.229$ for the 49 sessions rated $\le 1.5$ or $\ge 4.5$ against $0.208$ for the middle mass. Average rank error is thus similar at the extremes and in the middle, but the two level-1 sessions give little evidence about identifying the poorest conversations. The predictions stay near the center, the expected behavior of a regularized model trained on concentrated targets, and an argument for growing the pooled corpus at the tails rather than uniformly (only two sessions in the whole pool are rated $1$).

\section{Reproducibility, Resources, and AI Assistance}
\label{app:repro}

\paragraph{Compute.} All experiments ran on two university workstations with one NVIDIA RTX 4090 (24\,GB) each. LLM judging totals roughly 40 GPU-hours (195 sessions $\times$ 8 full judge runs plus the level-description control, screening and seed-stability reruns, 12B--30B models via ollama 0.32.1, 4-bit quantizations); feature extraction ran on the DISCOVER platform; the regression sweeps are CPU-only (scikit-learn \citealp{pedregosa2011scikit}; mord \citealp{pedregosa2017mord}; SciPy \citealp{virtanen2020scipy}) and complete in a few hours.

\paragraph{Determinism.} Every stochastic component is seeded, in the protocol and in the earlier exploratory sweeps outside it: cross-validation folds, mutual-information estimation, gradient boosting, and PCA. The last is easy to miss: for feature blocks wider than a few hundred columns scikit-learn's PCA switches to a randomized solver whose seed defaults to unset. The effect is modest but real: re-running the two wide blocks with the seed pinned moved 107 of their 300 projection cells, by $.01$ on average and up to $.10$, and it is confined to those blocks; every narrow block reproduces bit-identically either way. The per-judge seed-agreement figures reported in Appendix~\ref{app:mapping} are kept as a table (\texttt{seed\_agreement.csv}) rather than quoted only in prose; the per-session judge outputs behind them are cached, so the agreement report recomputes offline.

\paragraph{Provenance of every printed number.} The earlier exploratory sweeps outside the protocol live in six result CSVs, and nine feature sets were run twice: once on the original feature matrix and once after the attribution repair; a loader resolves each feature set to exactly one sweep (repaired build first) and asserts the resolution, so no cell can be an average of two experiments. The primary system, the in-domain reference, the bootstrap, the size- and test-matched controls and the appendix controls are one script family that refits from the feature matrices under the inductive protocol; every table in Section~\ref{sec:results} and Appendices~\ref{app:matchedn}--\ref{app:errors}, and Figure~\ref{fig:families}, is generated from its outputs. The code is available at \url{https://github.com/saveli/transfer-judges}: the feature and judging pipeline with the verbatim judge prompts, the fixed-configuration inductive fit, the environment lock and the ollama model digests, together with the out-of-fold predictions of the compact stack and the language-only system for every session under opaque per-domain indices, so the primary numbers can be recomputed from the file. The recordings and their session-level derivatives are not released, for the reasons given in Ethical Considerations.

\paragraph{Models and licenses.} Judge models: gemma3-12B (Gemma Terms of Use), qwen3-14B, qwen3.5-27B, qwen3.6-27B and granite4.1-8B/30B (Apache 2.0), gemma4-12B (Gemma Terms of Use), phi4-14B (MIT). Feature extractors: WhisperX, openSMILE/eGeMAPS (audEERING research license, non-commercial), EmotiEffLib, a wav2vec2 speech-emotion model \citep{wagner2023dawn}, and multilingual sentence/sentiment models \citep{reimers2019sbert,barbieri2022xlmt}; all used within their research license terms. No model was fine-tuned; all inference runs on institutional hardware.

\paragraph{Data.} The three corpora contain identifiable recordings of consenting students and actors and are not released, following the consent terms under which they were collected and the practice of the corpus publications (Section~\ref{sec:data}). Feature matrices and judge outputs are session-level derivatives of those recordings, so they are not released either. No recording, transcript, or personally identifying information leaves the institution's infrastructure at any stage; remote inference within the institution transfers transcripts transiently over encrypted channels without storing them.

\paragraph{AI assistance.} We used generative AI (Claude, Codex) for assistance in coding and in drafting text passages of this paper. Apart from the local judge models that are the object of study (Section~\ref{sec:judges}), no labels, targets, or reported results were AI-generated. All decisions and responsibilities rest with the authors.

\end{document}

%% file: primary_table.tex
\begin{table*}[t]
\centering\small
\setlength{\tabcolsep}{5pt}
\begin{tabular}{lcccc}
\toprule
Training setting & \textsc{Ticks} ($n{=}57$) & \textsc{Vacc} ($n{=}53$) & \textsc{Teach} ($n{=}85$) & mean \\
\midrule
\multicolumn{5}{l}{\emph{Compact stack (primary system): four judges $+$ compact nonverbal block and its deltas}}\\
cross-domain (training $n{=}138/142/110$) & .485 & .517 & .598 & \textbf{.533} \\
\;\;training size matched ($m{=}46/42/68$, 200 draws) & .486 & .426 & .607 & .506 \\
in-domain 5-fold CV, 20 splits (sd) & .348 (.048) & .380 (.054) & .533 (.030) & .420 \\
gap, paired bootstrap [95\% CI] & & & \multicolumn{2}{r}{$+.113$ $[+.067, +.157]$} \\
gap on identical test fifths (splits $>0$) & $+.098$ (19/20) & $+.070$ (19/20) & $+.025$ (18/20) & $+.065$ (19/20) \\
\midrule
\multicolumn{5}{l}{\emph{Language-only: the four LLM judges}}\\
cross-domain & .485 & .486 & .597 & .523 \\
\;\;training size matched & .476 & .452 & .595 & .508 \\
in-domain 5-fold CV, 20 splits (sd) & .415 (.026) & .423 (.031) & .559 (.024) & .466 \\
gap, paired bootstrap [95\% CI] & & & \multicolumn{2}{r}{$+.057$ $[+.038, +.077]$} \\
gap on identical test fifths (splits $>0$) & $+.014$ (16/20) & $-.005$ (7/20) & $+.007$ (15/20) & $+.006$ (14/20) \\
\bottomrule
\end{tabular}
\caption{Primary system under the inductive protocol: Spearman $\rho$ per
held-out domain and their mean, for both training settings on the same
sessions, with the training-size and identical-test-fifth controls and the
paired session bootstrap of Section~\ref{sec:protocol}. sd = across the 20
in-domain splits; $k/20$ = splits with a positive gap.}
\label{tab:primary}
\end{table*}

%% file: single_source_table.tex
\begin{table}[t]\centering\small
\setlength{\tabcolsep}{5pt}
\begin{tabular}{llccc}
\toprule
Training & Test & $n_\mathrm{train}$ & stack & language-only \\
\midrule
\textsc{Vacc} & \textsc{Ticks} & 53 & \textbf{.485} & .482 \\
\textsc{Teach} & \textsc{Ticks} & 85 & \textbf{.488} & .479 \\
\textsc{Ticks} & \textsc{Vacc} & 57 & .469 & \textbf{.492} \\
\textsc{Teach} & \textsc{Vacc} & 85 & \textbf{.507} & .487 \\
\textsc{Ticks} & \textsc{Teach} & 57 & \textbf{.631} & .587 \\
\textsc{Vacc} & \textsc{Teach} & 53 & \textbf{.608} & .589 \\
\bottomrule
\end{tabular}
\caption{Directed single-source transfer, fixed configuration (Spearman
$\rho$ on the full test corpus).}
\label{tab:single}
\end{table}

%% file: judge_table.tex
\begin{table}[t]
\centering\small
\begin{tabular}{@{}p{0.28\columnwidth}p{0.40\columnwidth}cc@{}}
\toprule
Battery & Model (family) & \#c & ens. \\
\midrule
generic dialogue & gemma3-12B (gemma) & 5 & $\bullet$ \\
generic dialogue & qwen3-14B, think (qwen) & 5 & $\bullet$ \\
BGR$+$OSCE & gemma3-12B (gemma) & 8 & $\bullet$ \\
\;\;anti-band wording & gemma3-12B (gemma) & 8 & \\
\;\;anti-band wording & qwen3.6-27B (qwen) & 8 & \\
\;\;anti-band wording & granite4.1-30B (granite) & 8 & $\bullet$ \\
\;\;anti-band wording & phi4-14B (phi) & 8 & \\
phase$+$technique & gemma3-12B (gemma) & 7 & \\
\bottomrule
\end{tabular}
\caption{The eight full judge runs: three construct batteries (one of them
in two promptings, Appendix~\ref{app:variants}) across five models from four
open-weight families \citep{gemma3,qwen3,granite2026,abdin2024phi}. \#c = constructs
returned per session; \emph{ens.} marks the four judges of the primary
ensemble, chosen by the inner selection of Section~\ref{sec:protocol}
(Appendix~\ref{app:matchedn}, Membership selection; picks in
Table~\ref{tab:nested}); qwen3.6, phi4 and the anti-band gemma3 run are
candidates the stack's inner selection does not pick; the phase$+$technique
judge was not a candidate (its battery overlaps BGR$+$OSCE). Each run scores
all 195 sessions. The two generic-dialogue runs share a prompt and differ
only in model, which is what makes them ensemble partners with decorrelated
errors.}
\label{tab:judges}
\end{table}

%% file: construct_table.tex
\begin{table}[t]
\centering\small
\begin{tabular}{lcccc}
\toprule
Construct (judge) & \textsc{Tic} & \textsc{Vac} & \textsc{Tea} & min \\
\midrule
phase completeness (P) & .33 & \textbf{.44} & \underline{.40} & \textbf{.33} \\
coherence (B) & .33 & \underline{.35} & .38 & \underline{.33} \\
overall grade (B) & .32 & .29 & \textbf{.41} & .29 \\
clarity (think) & .31 & .27 & .28 & .27 \\
structure (generic) & .26 & .31 & .27 & .26 \\
empathy (generic) & .30 & .25 & .38 & .25 \\
\midrule
BGR empathy (B) & .20 & $-$.05 & .23 & $-$.05 \\
structuring (P) & $-$.09 & .19 & .15 & $-$.09 \\
BGR verbal (B) & $-$.18 & .27 & .15 & $-$.18 \\
\bottomrule
\end{tabular}
\caption{Per-construct Spearman $\rho$ with the expert target, per domain
(B = BGR+OSCE judge, P = phase+technique judge, think = thinking variant);
the six most domain-consistent constructs (top, ranked by their weakest
domain) and the three least stable (bottom). Bold = best, underlined =
second best per column, over all constructs of all judges.}
\label{tab:constructs}
\end{table}

%% file: nested_table.tex
\begin{table}[ht]\centering\footnotesize\setlength{\tabcolsep}{3pt}
\begin{tabular}{llcccc}
\toprule
Membership & System & \textsc{Ticks} & \textsc{Vacc} & \textsc{Teach} & mean \\
\midrule
nested & compact stack & .485 & .517 & .598 & .533 \\
nested & language-only & .485 & .403 & .597 & .495 \\
nested & nv-dyadic stack & .464 & .425 & .575 & .488 \\
\midrule
primary & compact stack & .485 & .517 & .598 & .533 \\
primary & language-only & .485 & .486 & .597 & .523 \\
\bottomrule
\end{tabular}
\caption{Membership selection under LODO (Appendix~\ref{app:matchedn}).
\emph{nested}: membership chosen per outer fold by the inner rule. The compact
stack picks the primary membership in all three folds; language-only picks
it for \textsc{Ticks} and \textsc{Teach} and base three $+$ \{granite,
qwen3.6, phi4\} for \textsc{Vacc}; the nonverbal-dyadic stack picks a
different membership per fold. \emph{primary}: the four-judge membership
fixed for both systems.}
\label{tab:nested}
\end{table}

%% file: matchedn_table.tex
\begin{table*}[t]\centering\small
\begin{tabular}{llcccccc}
\toprule
System & Held-out & $n$ & $m$ & full & matched (sd) & in-domain (sd) & draws $>$ in-domain \\
\midrule
stack & \textsc{Ticks} & 57 & 46 & .485 & .486 (.022) & .348 (.048) & 100\% \\
stack & \textsc{Vacc} & 53 & 42 & .517 & .426 (.098) & .380 (.054) & 76\% \\
stack & \textsc{Teach} & 85 & 68 & .598 & .607 (.014) & .533 (.030) & 100\% \\
lang.-only & \textsc{Ticks} & 57 & 46 & .485 & .476 (.017) & .415 (.026) & 99\% \\
lang.-only & \textsc{Vacc} & 53 & 42 & .486 & .452 (.075) & .423 (.031) & 82\% \\
lang.-only & \textsc{Teach} & 85 & 68 & .597 & .595 (.005) & .559 (.024) & 100\% \\
\bottomrule
\end{tabular}
\caption{Size-matched cross-domain training against repeated in-domain CV,
per held-out domain. sd = across the 200 draws or the 20 splits.}
\label{tab:matchedn}
\end{table*}

%% file: main_table.tex
\begin{table*}[t]
\centering\small
\setlength{\tabcolsep}{4pt}
\begin{tabular}{lcccc|c}
\toprule
 & \multicolumn{4}{c|}{cross-domain $\rho$} & in-domain \\
Feature set & \textsc{Ticks} & \textsc{Vacc} & \textsc{Teach} & mean & mean \\
\midrule
\multicolumn{6}{l}{\emph{Language: LLM judges}}\\
four-judge ensemble (language-only) & .485 & .486 & .597 & .523 & \textbf{.466} \\
granite4.1-30B judge & .440 & .329 & .575 & .448 & .353 \\
BGR+OSCE judge (gemma3) & .398 & .434 & .423 & .418 & .283 \\
qwen3.6 judge (3-seed mean) & .485 & .279 & .363 & .375 & .274 \\
phase+technique judge & .371 & .369 & .343 & .361 & .223 \\
generic dialogue judge (gemma3) & .340 & .324 & .299 & .321 & .084 \\
generic dialogue judge (qwen3-14B think) & .370 & .265 & .389 & .341 & .190 \\
\midrule
\multicolumn{6}{l}{\emph{Language $+$ nonverbal}}\\
compact stack (primary system) & .485 & .517 & .598 & \textbf{.533} & .420 \\
\midrule
\multicolumn{6}{l}{\emph{Nonverbal and surface baselines}}\\
compact$+$deltas & .249 & .100 & .134 & .161 & $-$.066 \\
nonverbal-dyadic & $-$.026 & .000 & .037 & .004 & .022 \\
shallow transcript stats & .000 & .032 & .101 & .044 & $-$.054 \\
sentence embeddings & .087 & .001 & .067 & .052 & .141 \\
\bottomrule
\end{tabular}
\caption{Per-fold values behind Figure~\ref{fig:families}: feature families
at the fixed configuration (PCA-10 $+$ elastic net) under the inductive
protocol, cross-domain Spearman $\rho$ per held-out domain and mean, and the
20-split in-domain CV mean.}
\label{tab:main}
\end{table*}

%% file: calib_table.tex
\begin{table}[ht]\centering\small
\begin{tabular}{lccccc}
\toprule
 & $\rho$ & MAE & mean-only & QWK & sd(pred) \\
\midrule
\textsc{Ticks} & .485 & .741 & .908 & .381 & 0.45 \\
\textsc{Vacc} & .517 & .751 & .816 & .215 & 0.42 \\
\textsc{Teach} & .598 & .645 & .754 & .343 & 0.32 \\
\midrule
pooled & .533 & .702 & .815 & .327 & 0.44 \\
\bottomrule
\end{tabular}
\caption{Absolute-error and agreement metrics for the primary system, per
domain and pooled; $\rho$ pooled = mean over domains; MAE, mean-only, QWK
and sd(pred) pooled over the 195 sessions. \emph{mean-only} is the MAE of a
constant at the evaluated sessions' own target mean (per domain, or pooled),
an oracle; QWK = quadratic-weighted $\kappa$ against the
rounded target.}
\label{tab:calib}
\end{table}